\def\year{2022}\relax
\documentclass[letterpaper]{article} 
\usepackage{aaai22}  
\usepackage{times}  
\usepackage{helvet}  
\usepackage{courier}  
\usepackage[hyphens]{url}  
\usepackage{graphicx} 
\usepackage{natbib}  
\usepackage{caption} 
\DeclareCaptionStyle{ruled}{labelfont=normalfont,labelsep=colon,strut=off} 
\usepackage[utf8]{inputenc} 
\usepackage[T1]{fontenc}    
\usepackage{url}            
\usepackage{booktabs}       
\usepackage{amsfonts}       
\usepackage{nicefrac}       
\usepackage{microtype}      

\usepackage{algorithm}
\usepackage{algorithmic}

\usepackage{amsmath}
\usepackage{url}
\usepackage{graphicx}  
\usepackage{color}
\usepackage{subfigure}

\usepackage{multirow}
\usepackage{amssymb}
\usepackage{pifont}
\newcommand{\cmark}{\ding{51}}%
\usepackage{newfloat}
\usepackage{listings}
\floatstyle{ruled}
\newfloat{listing}{tb}{lst}{}
\floatname{listing}{Listing}
\title{Image-Adaptive YOLO for Object Detection in Adverse  Weather Conditions}
\author {
	Wenyu Liu\textsuperscript{\rm 1,2}\thanks{This work was done when the author was visiting Alibaba as a research intern.},
	Gaofeng Ren \textsuperscript{\rm 3},
	Runsheng Yu \textsuperscript{\rm 4},
	Shi Guo \textsuperscript{\rm 5},
	Jianke Zhu \textsuperscript{\rm 1,2 \footnote{corresponding author}},
	Lei Zhang \textsuperscript{\rm 3,5}
}
\affiliations {
	\textsuperscript{\rm 1} College of Computer Science and Technology, Zhejiang University\\
	\textsuperscript{\rm 2} Alibaba-Zhejiang University Joint Institute of Frontier Technologies\\

	\textsuperscript{\rm 3} DAMO Academy, Alibaba Group\\
	\textsuperscript{\rm 4} The Hong Kong University of Science and Technology\\

	\textsuperscript{\rm 5} The HongKong Polytechnic University\\

	\{liuwenyu.lwy, jkzhu\}@zju.edu.cn, \{gaof.ren, runshengyu\}@gmail.com, \{csshiguo, cslzhang\}@comp.polyu.edu.hk}
	
\usepackage{bibentry}

\begin{document}

\maketitle

\begin{abstract}
Though deep learning-based object detection methods have achieved promising results on the conventional datasets, it is still challenging to locate objects from the low-quality images captured in adverse weather conditions. The existing methods either have difficulties in balancing the tasks of image enhancement and object detection, or often ignore the latent information beneficial for detection. To alleviate this problem, we propose a novel Image-Adaptive YOLO (IA-YOLO) framework, where each image can be adaptively enhanced for better detection performance. Specifically, a differentiable image processing (DIP) module is presented to take into account the adverse weather conditions for YOLO detector, whose parameters are predicted by a small convolutional neural network (CNN-PP). We learn CNN-PP and YOLOv3 jointly in an end-to-end fashion, which ensures that CNN-PP can learn an appropriate DIP to enhance the image for detection in a weakly supervised manner. Our proposed IA-YOLO approach can adaptively process images in both normal and adverse weather conditions. The experimental results are very encouraging, demonstrating the effectiveness of our proposed IA-YOLO method in both foggy and low-light scenarios. The source code can be found at https://github.com/wenyyu/Image-Adaptive-YOLO. 


\end{abstract}

\section{Introduction}
\begin{figure}[t]
\centering
    \subfigure[\scriptsize{ YOLO \uppercase\expandafter{\romannumeral2} (Baseline)}]{
    \includegraphics[width=0.22\textwidth]{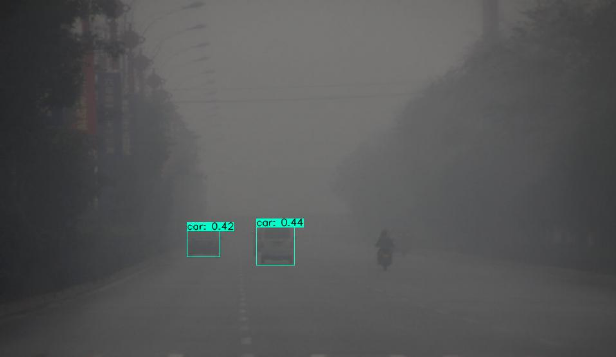}
    }
    \subfigure[\scriptsize{IA-YOLO (Ours)}]{
    \includegraphics[width=0.22\textwidth]{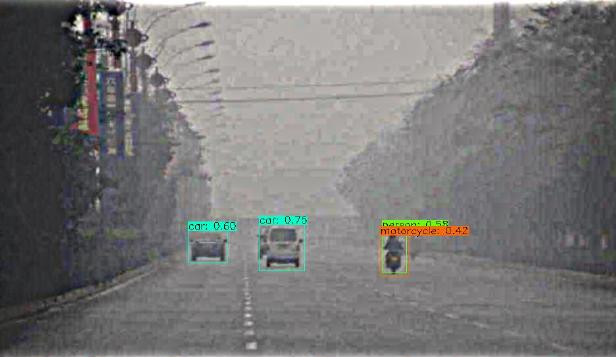}
    }
    \caption{In the real-world foggy condition, our method can adaptively output clearer images with sharper edges around objects' boundary, and consequently produce higher confidence detection results with fewer missing instances.}
    \label{fig1}
\end{figure}

CNN-based methods have become prevailed in object detection~\citep{ren2015faster, redmon2018yolov3}. They not only have achieved promising performance on benchmark datasets~\citep{deng2009imagenet, everingham2010pascal,lin2014microsoft} but also have been deployed in the real-world applications such as autonomous driving \cite{wang2019pso}. Due to the domain shift in input images~\cite{sindagi2020prior},  general object detection models trained by high quality images often fail to achieve satisfactory results under adverse weather conditions (e.g., foggy and dark light). \citet{narasimhan2002vision} and \citet{you2015adherent} suggested that an image captured under adverse weather can be decomposed into a clean image and its corresponding weather-specific information, and the image quality degradation in adverse weather is mainly caused by the interaction between weather-specific information and objects, which leads to poor detection performance. Fig.~\ref{fig1} shows an example of object detection in foggy conditions. One can see that if the image can be properly enhanced with respect to the weather condition, more latent information about the original blurred objects and the misidentified objects can be recovered.



To tackle this challenging problem, \citet{huang2020dsnet} employed two subnetworks to jointly learn visibility enhancement and object detection, where the impact of image degradation is reduced by sharing the feature extraction layers. However, it is hard to tune the parameters to balance the weights between detection and restoration during training. Another approach is to dilute the influence of weather-specific information by preprocessing the image with existing methods such as image defogging~\citep{MSBDN-DFF, liuICCV2019GridDehazeNet} and image enhancement~\cite{Zero-DCE}. However, complicated image restoration networks have to be included in these methods, which need to be trained separately with pixel-level supervision. This requires to manually label the images for restoration. This problem can also be treated as an unsupervised domain adaptation task~\citep{chen2018domain, hnewa2021multiscale}. Compared with training detectors with clear images (source image), it is assumed that images captured under adverse weathers (target image) have a distribution shift. These methods mostly adopt the domain adaptation principles and focus on aligning the features of two distributions, and the latent information which can be obtained in the process of weather-based image restoration is usually ignored.

To address the above limitations, we propose an ingenious image-adaptive object detection method, called IA-YOLO. Specifically, we suggest a fully differentiable image processing module (DIP), whose hyperparameters are adaptively learned by a small CNN-based parameter predictor (CNN-PP). The CNN-PP adaptively predicts the DIP's hyperparameters according to the brightness, color, tone and weather-specific information of the input image. After processed by the DIP module, the interference of weather-specific information in the image can be supressed while the latent information can be restored. We present a joint optimization scheme to learn the DIP, CNN-PP and YOLOv3 backbone detection network~\cite{redmon2018yolov3} in an end-to-end manner. To enhance the image for detection, CNN-PP is weakly supervised to learn an appropriate DIP through the bounding box annotations. Additionally, we make use of the images in both normal and adverse weather conditions to train the proposed network. By taking advantages of the CNN-PP network, our proposed IA-YOLO approach is able to adaptively deal with images affected by different degrees of weather conditions. Fig.~\ref{fig1} shows an example of the detection result by our proposed method.



 

The highlights of this work are: 1) an image-adaptive detection framework is proposed, which achieves promising performance in both normal and adverse weather conditions; 2) a white-box differentiable image processing module is proposed, whose hyperparameters are predicted by a weakly supervised parameter prediction network; 3) encouraging experimental results are achieved on both synthetic testbeds (VOC\_Foggy and VOC\_Dark) and real-world datasets (RTTS and ExDark), comparing against the previous methods.

\begin{figure*}[t]
\centering
\includegraphics[width=0.9\textwidth]{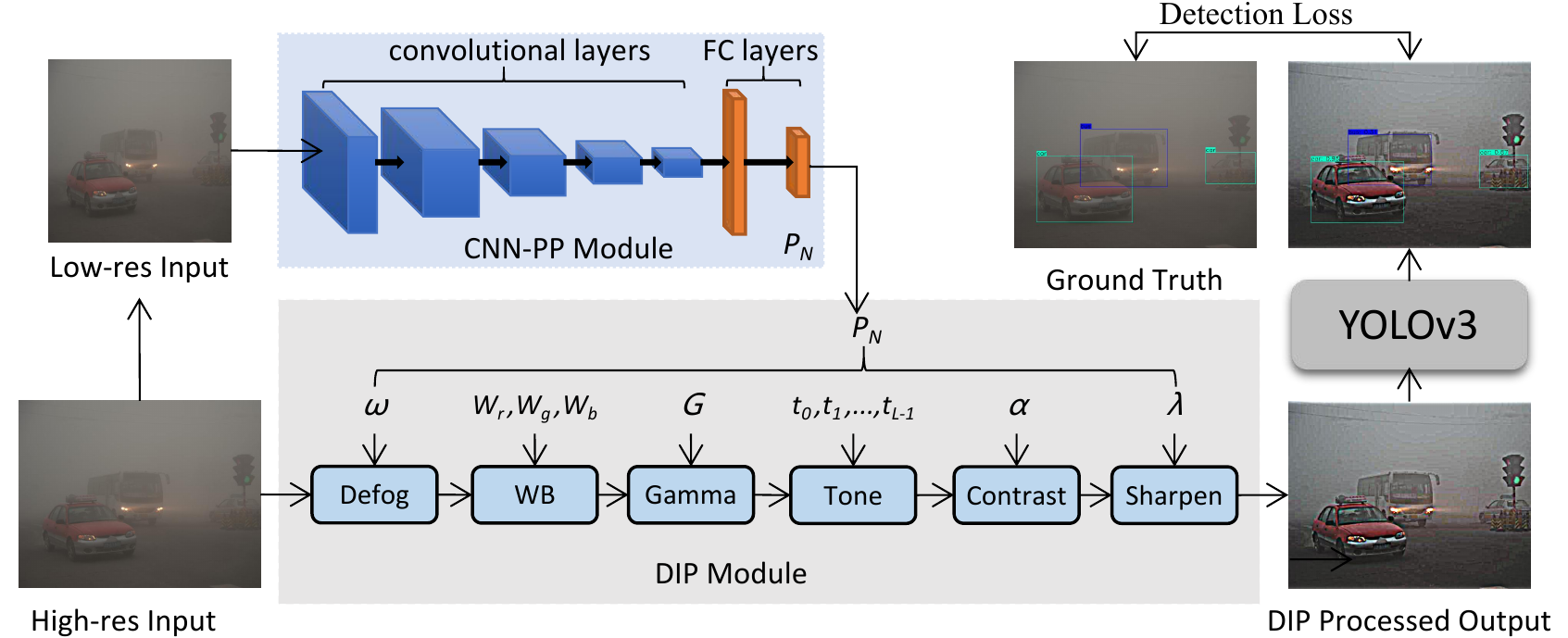} 
\caption{The end-to-end training pipeline of the proposed IA-YOLO framework. Our method learns a YOLO with a small CNN-based parameter predictor (CNN-PP), which employs the downsampled input image to predict the hyperparamters of filters in the DIP module. The high-resolution input images are processed by DIP's filters to aid YOLOv3 achieve high detection performance. The \emph{Defog filter} is only used in foggy conditions.}
\label{fig2}
\end{figure*}

\section{Related Work}

\subsection{Object Detection} As a fundamental task in computer vision, object detection has received intensive attentions. The object detection methods can be roughly divided into two categories~\citep{zhao2019object}. One category is the region proposal-based methods~\citep{girshick2014rich, girshick2015fast, ren2015faster}, which first generate regions of interest (RoIs) from the image, and then classify them by training neural networks. Another category is one-stage regression-based approaches such as the YOLO series~\citep{redmon2016you,  redmon2017yolo9000, redmon2018yolov3, bochkovskiy2020yolov4} and SSD~\citep{liu2016ssd}, where the object labels and bounding box coordinates are predicted by a single CNN. In this paper, we employ the classic one-stage detector YOLOv3~\cite{redmon2018yolov3} as the baseline detector, and improve its performance in adverse conditions.


\subsection{Image Adaptation}
Image adaptation is widely used in image enhancement. To appropriately enhance an image, some traditional methods~\citep{polesel2000image, yu2004fast, wang2021adaptive} adaptively calculate the parameters of image transformation according to the corresponding image features. For example,~\citet{wang2021adaptive} proposed an brightness adjustment function that adaptively adjusts the enhancement parameters based on the illumination distribution characteristics of an input image. 


To achieve adaptive image enhancement, ~\citep{hu2018exposure,yu2018deepexposure,zeng2020learning} employed a small CNN to flexibly learn the hyperparameters of image transformation. \citet{hu2018exposure} proposed a post-processing framework with a set of differentiable filters, where deep reinforcement learning (DRL) is used to generate the image operation and filter parameters according to the quality of the current retouched image. \citet{zeng2020learning} utilized a small CNN to learn image-adaptive 3D LUTs according to the global context such as brightness, color and tones.


\subsection{Object Detection in Adverse Conditions}

Compared against general object detection, few research efforts have been made on object detection under adverse weather conditions. A straightforward approach is to preprocess the image by using the classical dehazing or image enhancement methods~\citep{Zero-DCE, he2009single, liuICCV2019GridDehazeNet, MSBDN-DFF, qin2020ffa}, which are originally designed to remove fog and enhance image quality. However, the improvement on image quality may not definitely benefit the detection performance. Some prior-based methods~\citep{li2017aod, huang2020dsnet} jointly perform image enhancement and detection to attenuate the interference of adverse weather-specific information. \citet{sindagi2020prior} proposed an unsupervised prior-based domain adversarial object detection framework for detection in hazy and rainy conditions. A few methods~\citep{chen2018domain, zhang2021domain, hnewa2021multiscale} make use of domain adaptation to address this problem. \citet{hnewa2021multiscale} assumed that there is domain shift between images captured under normal and adverse weather conditions. They designed a multi-scale domain adaptive YOLO that supports domain adaptation in different layers at the feature extraction stage.

\section{Proposed Method}

The images captured in adverse weather conditions have poor visibility due to the interference of weather-specific information, causing difficulties in object detection. To tackle this challenge, we suggest an image-adaptive detection framework by removing weather-specific information and revealing more latent information. As illustrated in Fig.~\ref{fig2}, the whole pipeline consists of a CNN-based parameter predictor (CNN-PP), a differentiable image processing module (DIP) and a detection network. We first resize the input image into the size of $256\times 256$, and feed it into CNN-PP to predict DIP's parameters. Then, the image filtered by DIP module is treated as the input for YOLOv3 detector. We present an end-to-end hybrid data training scheme with detection loss so that the CNN-PP can learn an appropriate DIP to enhance the image for object detection in a weakly supervised manner. 

\subsection{DIP Module}

As in \cite{hu2018exposure}, the design of image filters should conform to the principle of differentiability and resolution-independent. For the gradient-based optimization of CNN-PP, the filters should be differentiable to allow training the networks by backpropagation. Since CNN will consume a lot of computing resources to process high resolution images (e.g., $4000 \times 3000$), in this paper we learn the filter parameters from the downsampled low-resolution image of size $256\times 256$, and then apply the same filter to the image of original resolution. Therefore, these filters need to be independent of image resolution.


Our proposed DIP module consists of six differentiable filters with adjustable hyperparameters, including \emph{Defog}, \emph{White Balance(WB)}, \emph{Gamma}, \emph{Contrast}, \emph{Tone} and \emph{Sharpen}. As in~\citep{hu2018exposure}, the standard color and tone operators, such as \emph{WB}, \emph{Gamma}, \emph{Contrast} and \emph{Tone}, can be expressed as pixel-wise filters. Therefore, our designed filters can be categorized into \emph{Defog}, \emph{Pixel-wise filters} and \emph{Sharpen}. Among these filters, \emph{Defog filter} is specially designed for foggy scenes. The details are as follows.

\subsubsection{Pixel-wise Filters.} The pixel-wise filters map an input pixel value $P_{i} = (r_{i}, g_{i}, b_{i})$ into an output pixel value $P_{o} = (r_{o}, g_{o}, b_{o})$, in which $(r, g, b)$ represent the values of the three color channels red, green and blue, respectively. The mapping functions of the four pixel-wise filters are listed in Table \ref{table1}, where the second column lists the parameters to be optimized in our approach. \emph{WB} and \emph{Gamma} are simple multiplication and power transformations. Obviously, their mapping functions are differentiable with respect to both the input image and the parameters. 

\begin{table}[t]
\centering
\scalebox{0.81}{
\begin{tabular}{lll}
    \toprule
    Filter & Parameters & Mapping Function  \\
    \midrule
    WB & $W_{r}, W_{g}, W_{b}$: factors & $P_{o} = (W_{r}r_{i}, W_{g}g_{i}, W_{b}b_{i})$ \\
    Gamma & $G$: gamma value & $P_{o} = P_{i}^{G}$  \\
    Contrast & $\alpha $: contrast value & $P_{o} = \alpha \cdot En(P_{i})+(1-\alpha) \cdot P_{i}$ \\
    Tone & $t_{i}$: tone params & $P_{o} = (L_{t_{r}}(r_{i}), L_{t_{g}}(g_{i}), L_{t_{b}}(b_{i}))$ \\
    \bottomrule
\end{tabular}
}
\caption{The mapping functions of pixel-wise filters.}
\label{table1}
\end{table}

The differentiable contrast filters are designed with an input parameter to set the linear interpolation between the original image and the fully enhanced image. As shown in Table~\ref{table1}, the definition of $En(P_{i})$ in the mapping function is as follows:
\begin{equation}
Lum(P_{i}) = 0.27r_{i}+0.67g_{i}+0.06b_{i} \label{1}
\end{equation}
\vspace{-0.15in}
\begin{equation}
EnLum(P_{i})=\frac{1}{2}(1-\cos(\pi \times (Lum(P_{i}))))\\\label{2}
\end{equation}
\begin{equation}
En(P_{i})=P_{i}\times\frac{EnLum(P_{i})}{Lum(P_{i})}\label{3}
\end{equation}


As in~\cite{hu2018exposure}, we design the tone filter as a monotonic and piecewise-linear function. We learn tone filter with \emph{L} parameters, represented as $\{t_{0},t_{1},\dots,t_{L-1}\}$. The points of tone curve are denoted as $(k/L,T_{k}/T_{L})$, in which $T_{k}= \sum_{i=0}^{k-1}t_{l}$. Furthermore, the mapping function is expressed by the differentiable parameters, which enable the function to be differentiable with respect to both the input image and the parameters $\{t_{0},t_{1},\dots,t_{L-1}\}$ as below
\begin{equation}
P_{o} =\frac{1}{T_{L}}\sum_{j=0}^{L-1}clip(L\cdot P_{i}-j,0,1)t_{k}\label{4}
\end{equation}

\subsubsection{Sharpen Filter.} Image sharpening can highlight the image details. Like the unsharpen mask technique~\citep{polesel2000image}, the sharpening process can be described as follows:
\begin{equation}
F(x, \lambda) = I(x) + \lambda(I(x)- Gau(I(x)))\label{5}\\
\end{equation} 
where $I(x)$ is the input image, $Gau(I(x))$ denotes Gaussian filter, and $\lambda$ is a positive scaling factor. This sharpening operation is differentiable with respect to both $x$ and $\lambda$. Note that the sharpening degree can be tuned for better object detection performance by optimizing $\lambda$.

\subsubsection{Defog Filter.} Motivated by the dark channel prior method~\cite{he2009single}, we design a defog filter with a learnable parameter. Based on the atmospheric scattering model~\citep{mccartney1976optics,narasimhan2002vision}, the formation of a hazy image can be formulated as follows:
\begin{equation}
I(x) = J(x)t(x) + A(1-t(x))\label{6}
\end{equation} 
where $I(x)$ is the foggy image, and $J(x)$ represents the scene radiance (clean image). $A$ is the global atmospheric light, and $t(x)$ is the medium transmission map, which is defined as:
\begin{equation}
t(x) = e ^{-\beta} d(x)\label{7}
\end{equation}
where $\beta$ represents the scattering coefficient of the atmosphere, and $d(x)$ is the scene depth.

In order to recover the clean image $J(x)$, the key is to obtain the atmospheric light $A$ and the transmission map $t(x)$. To this end, we first compute the dark channel map of the haze image $I(x)$, and pick the top 1000 brightest pixels. Then, $A$ is estimated by averaging those 1000 pixels of the corresponding position of the haze image $I(x)$. 

According to Eq.~(\ref{6}), one can derive an approximate solution of $t(x)$ as follows ~\citep{he2009single}
\begin{equation}
t(x) = 1- \min \limits_{C} \left( \min \limits_{y \in \Omega(x)} \frac{I^C (y)}{A^C}  \right) \label{8}
\end{equation}
We further introduce a parameter $\omega$ to control the degree of defogging as follows:
\begin{equation}
t(x, \omega) = 1-  \omega \min \limits_{C} \left (\min \limits_{y \in \Omega(x)} \frac{I^C (y)}{A^C}\right) \label{9}
\end{equation}
Since the above operation is differentiable, we can optimize $\omega$ through back propagation to make \emph{defog filter} more conducive to foggy image detection.





\subsection{CNN-PP Module}
In camera image signal processing (ISP) pipeline, some adjustable filters are usually employed for image enhancement, whose hyperparameters are manually tuned by experienced engineers through visual inspection~\citep{mosleh2020hardware}. Generally, such a tuning process is very awkward and expensive to find suitable parameters for a broad range of scenes. To address this limitation, we propose to employ a small CNN as a parameter predictor to estimate the hyperparameters, which is very efficient.

Taking foggy scene as an example, the purpose of CNN-PP is to predict the DIP's parameters by understanding the global content of the image, such as brightness, color and tone, as well as the degree of fog. Therefore, the down-sampled image is sufficient to estimate these information, which can greatly save the computational cost. Given an input image of any resolution, we simply use bilinear interpolation to downsample it to $256 \times 256$ resolution. As shown in Fig.~\ref{fig2}, the CNN-PP network is composed of five convolutional blocks and two fully-connected layers. Each convolutional block includes a $3 \times 3$ convolutional layer with stride 2 and a leaky Relu. The final fully-connected layer outputs the hyperparameters for the DIP module. The output channel of these five convolutional layers are 16, 32, 32, 32 and 32, respectively. The CNN-PP model contains only 165K parameters when the total number of parameter is 15.

\subsection{Detection Network Module}
In this paper, we choose the one-stage detector YOLOv3 as the detection network, which is widely used in practical applications, including image editing, security monitoring, crowd detection and autonomous driving~\cite{zhang2021domain}. Compared with the previous version, YOLOv3 designs darknet-53 that is composed of successive $3 \times 3$ and $1 \times 1$ convolutional layers based on the idea of Resnet~\citep{he2016deep}. It implements multi-scale training by making predictions on multi-scale feature maps, so as to further improve the detection accuracy, especially for small objects. We adopt the same network architecture and loss functions as the original YOLOv3~\cite{redmon2018yolov3}.

\subsection{Hybrid Data Training}
In order to achieve ideal detection performance in both normal and adverse weather conditions, we adopt a hybrid data training scheme for IA-YOLO. Algorithm \ref{alg1} summarizes the training process of our proposed method. Each image has a probability of 2/3 to be randomly added with some kind of fog or be transformed to a low-light image before being input to the network for training. With both normal and synthetic low-quality training data, the whole pipeline is trained end-to-end with YOLOv3 detection loss, which ensures all modules in IA-YOLO can adapt to each other. Therefore, the CNN-PP module is weakly supervised by the detection loss without manually labeling ground truth images. The hybrid data training mode ensures that IA-YOLO can adaptively process the image according to the content of each image, so as to achieve high detection performance.



\begin{algorithm}[t]
  \caption{Image-Adaptive YOLO training procedure}
  \label{alg1}
  \begin{algorithmic}
\STATE Initialize the CNN-PP network $P ^\theta $ and the YOLOv3 network $D^ \beta$ with random weights $\theta $ and $\beta$.
\STATE Set the training stage: $num\_epochs= 80$, $batch\_size = 6$.
\STATE Prepare the normal dataset VOC\_norm\_trainval.
 \FOR{$i$ in $num\_epochs$ }
  \REPEAT
   \STATE Take a batch images $M$ from VOC\_norm\_trainval;
   \FOR{$j$ in $batch\_size$ } 
     \IF{$random.randint(0, 2) > 0$}
        \STATE Generate the foggy image $M(j)$ (Eq. (\ref{10}, \ref{11})), where $A=0.5$, $k=random.randint(0, 9), \beta = 0.01 \ast k + 0.05$  //for foggy conditions
        \STATE Generate the low-light image $M(j)$ by $f(x) = x^\gamma$,  where $\gamma = random.uniform(1.5, 5)$ //for low-light conditions
     \ENDIF
   \ENDFOR
    \STATE Compute DIP params by $P_{N}=P ^\theta (image\_batch) $;
    \STATE Perform DIP filter processing by $image\_batch = DIP(image\_batch, P_{N} )$;  
    \STATE Send $image\_batch$ to YOLOv3 network $D^ \beta$;
    \STATE  Update the CNN-PP network $P ^\theta $ and the YOLOv3 network $D^ \beta$ according to the YOLOv3 detection loss.
  \UNTIL{all images have been fed into training models}
 \ENDFOR
  \end{algorithmic}
\end{algorithm}

\section{Experiments}
We evaluate the effectiveness of our method in the scenarios of fog and low-light. The filter combination is [\emph{Defog, White Balance(WB), Gamma, Contrast, Tone, Shapen}], while the \emph{Defog filter} is only used in foggy conditions.

\subsection{Implementation Details}
We adopt the training protocol of \cite{redmon2018yolov3} in our proposed IA-YOLO approach. The backbone network for all experiments is Darknet-53. During training, We randomly resize the image to ($32N\times32N$), where $N$ $\in$ [9, 19]. Moreover, the data augmentation methods like image flipping, cropping and transformation are applied to expand the training dataset. Our IA-YOLO model is trained by the Adam optimizer~\citep{kingma2014adam} with 80 epochs. The starting learning rate is $10^{-4}$ and the batch size is 6. IA-YOLO predicts the bounding boxes at three different scales, and three anchors at each scale. We use Tensorflow for our experiments and run it on the Tesla V100 GPU.

\subsection{Experiments on Foggy Images}
\subsubsection{Datasets} There are few publicly available datasets for object detection in adverse weather conditions, and the data amounts are usually small for training a stable CNN-based detector. To facilitate fair comparison, we build upon the classic VOC dataset~\citep{everingham2010pascal} a VOC\_Foggy dataset according to the atmospheric scattering model~\cite{narasimhan2002vision}. Moreover, RTTS~\cite{li2018benchmarking} is a relatively comprehensive real-world dataset available in foggy conditions, which has 4,322 natural hazy images with five annotated object classes, namely, person, bicycle, car, bus and motorcycle. To form our training dataset, we select the data containing these five categories to add haze.




For VOC2007\_trainval and VOC2012\_trainval, we filtrate the images which include the above five classes of objects to build VOC\_norm\_trainval. VOC\_norm\_test is selected from the VOC2007\_test in a similar way. We also evaluate our method on RTTS. The statistics of the datasets are summarized into Table \ref{table2}.

\begin{table}[t]
\centering
\scalebox{0.85}{
\begin{tabular}{cccccccc}
    \toprule
    Dataset & image & ps & bc & car & bus & mc & Total \\
    \midrule
    V\_n\_tv & 8111 & 13256 & 1064 & 3267 & 822 & 1052 &19561 \\
    V\_n\_ts  & 2734 & 4528 & 337 & 1201 & 213 & 325 &6604 \\
    RTTS & 4322 & 7950 & 534 & 18413 & 1838 & 862 &29577 \\
    \bottomrule
\end{tabular}
}
\caption{Statistics of the used datasets, including V\_n\_tv (VOC\_norm\_trainval), V\_n\_ts (VOC\_norm\_test) and RTTS. The classes are ps (person), bc (bicycle), car, bus, and mc (motorcycle).}
\label{table2}
\end{table}

To avoid the computational cost on generating foggy images during the training process, we construct the VOC\_Foggy dataset offline. According to Eqs. (\ref{6}, \ref{7}), the foggy image $I(x)$ is obtained by
\begin{equation}
I(x) = J(x)e ^{-\beta} d(x) + A(1-e ^{-\beta} d(x))
\label{10}
\end{equation} $d(x)$ is defined as below:
\begin{equation}
d(x) = -0.04 \ast \rho + \sqrt{\max(row, col)}
\label{11}
\end{equation} 
 where $\rho$ is the Euclidean distance from the current pixel to the central pixel, \emph{row} and \emph{col} are the number of rows and columns of the image, respectively. By setting $A = 0.5$ and $\beta = 0.01 \ast i + 0.05$, where $i$ is a integer number from 0 to 9, ten different levels of fog can be added to each image. Based on the VOC\_norm\_trainval dataset, we generate the VOC\_Foggy\_trainval dataset ten times larger offline. To obtain the VOC\_Foggy\_test dataset, every image in VOC\_norm\_test 
is randomly processed with fog.

\begin{figure*}[t]
\centering
\includegraphics[width=1.0\textwidth]{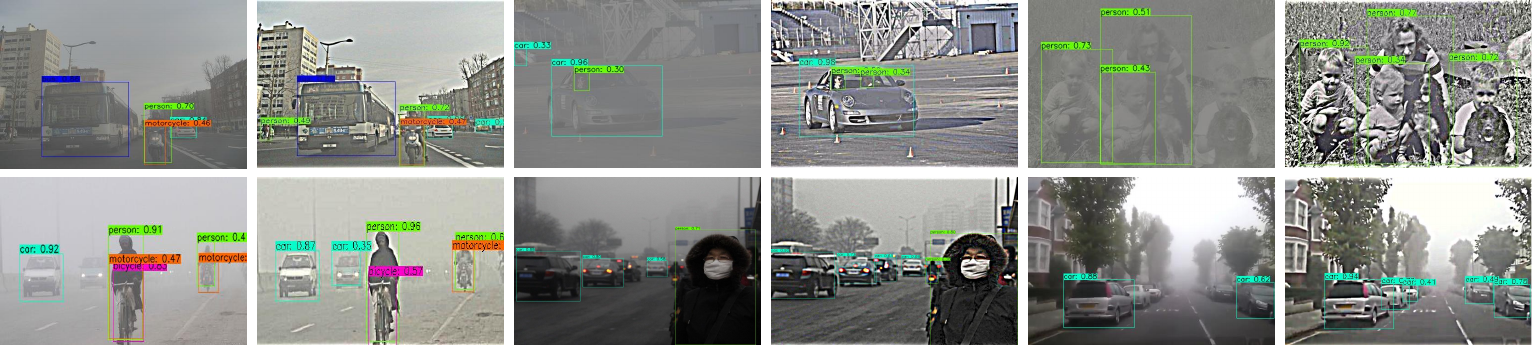} 
\caption{Detection results of YOLOv3 \uppercase\expandafter{\romannumeral2} (columns 1, 3 and 5) and our IA-YOLO (columns 2, 4 and 6) on synthetic VOC\_Foggy\_test images (top row) and real-world RTTS foggy images (bottom row). The proposed method learns to reduce the haze and sharpen image edge, which has better detection performance with fewer missed and wrong detections. }
\label{fig3}
\end{figure*}
\subsubsection{Experimental Results} To demonstrate the effectiveness of IA-YOLO, we compare our method with baseline YOLOv3, $Defog+Detect$, domain adaption~\cite{hnewa2021multiscale}, and multi-task learning~\cite{huang2020dsnet} on the three testing datasets. For $Defog+Detect$, we employ the defogging method as the preprocessing step and use YOLOv3 trained on VOC\_norm for detection. We select MSBDN~\cite{MSBDN-DFF} and GridDehaze~\cite{liuICCV2019GridDehazeNet} as the perprocessing methods, which are popular CNN-based dehazing methods. For domain adaption approach, we employ the one-stage multi-scale domain-adaptive detector DAYOLO~\cite{hnewa2021multiscale} with multiple domain adaptation paths and the corresponding domain classifiers at different scales of YOLOv3. For multi-task learning algorithms, we choose DSNet~\cite{huang2020dsnet} that jointly learns dehazing and detection in adverse weather conditions. We reproduce its detection subnetwork and restoration module by sharing the first five convolution layers of Yolov3, and jointly train two networks with hybrid data.

\begin{table}[t]
\centering
\scalebox{0.85}{
\begin{tabular}{l|l|l|ccc}

\hline
\multicolumn{2}{c|}{Method}&Train data & V\_n\_ts&V\_F\_t&RTTS\\
\cline{1-6}
  \multirow{2}{*}{Baseline} & YOLOv3 \uppercase\expandafter{\romannumeral1}&VOC\_norm&70.10 & 31.05 &28.82\\ 
  &YOLOv3 \uppercase\expandafter{\romannumeral2}&Hybrid data&64.13 & 63.40 &30.80\\
  \cline{1-3}
    \multirow{2}{*}{Defog} & MSBDN&VOC\_norm&/ & 57.38 &30.20\\ 
  &GridDehaze&VOC\_norm&/ & 58.23 &31.42\\
  \cline{1-3}
  DA& DAYOLO&Hybrid data&56.51 & 55.11 &29.93\\ \cline{1-3}
  Mul-task&DSNet&Hybrid data&53.29 & 67.40 &28.91\\ \cline{1-3}
  Ours&IA-YOLO&Hybrid data&\textbf{73.23} & \textbf{72.03} &\textbf{37.08}\\ 
  \hline
\end{tabular}
}
\caption{Performance comparison on foggy images. "DA" means Domain Adaption. The right three columns list the mAP on three test datasets, including V\_n\_ts (VOC\_norm\_test), V\_F\_t (VOC\_Foggy\_test) and RTTS.}
\label{table3}
\end{table}

Table \ref{table3} compares the mean average precision (mAP) results~\cite{pascal-voc-2012} between IA-YOLO and other competing algorithms in both the normal and hazy conditions. The second column lists the training data for each method, where "Hybrid data" means the hybrid data training scheme used in our proposed IA-YOLO. Comparing to the baseline (YOLO \uppercase\expandafter{\romannumeral1}), all methods have improvements on both synthetic and real-word foggy weather testing datasets, while only our IA-YOLO has no deterioration in the case of normal scenarios. This is because the previous methods are mainly designed to deal with object detection in foggy weather condition while sacrificing their performance on the normal weather images. For our proposed IA-YOLO method, the CNN-PP and DIP modules are able to adaptively process the image with different degrees of fog for object detection. As a result, our proposed IA-YOLO approach outperforms all the competing methods on the three testing datasets by a large margin, demonstrating its effectiveness on object detection in adverse weather conditions. 

Fig.~\ref{fig3} shows several visual examples of our IA-YOLO method and the baseline YOLOv3 \uppercase\expandafter{\romannumeral2}. Though for some instances, our adaptive DIP module produces some unwilling noises for visual perception, it greatly boosts the local image gradients based on image semantics, and leads to better detection performance.

\begin{figure}[t]
\centering
\includegraphics[width=0.45\textwidth]{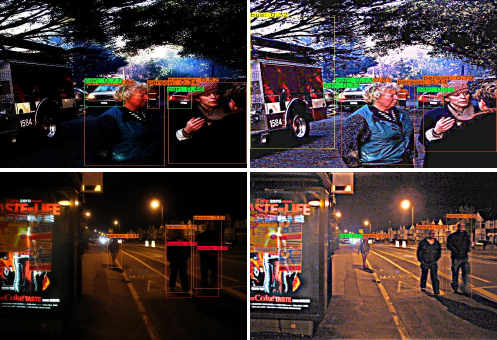} 
\caption{Detection results of YOLOv3 \uppercase\expandafter{\romannumeral2} (left column) and our IA-YOLO (right column) on synthetic VOC\_Dark\_test images (top row) and real-world ExDark low-light images (bottom row). The proposed method learns to enhance the image contrast with more details.}
\label{fig4}
\end{figure}

\subsection{Experiments on Low-light Images}
\subsubsection{Datasets} PSCAL VOC~\cite{everingham2010pascal} and the relatively comprehensive low-light detection dataset ExDark~\cite{loh2019getting} both contain ten categories of objects: \emph{Bicycle, Boat, Bottle, Bus, Car, Cat, Chair, Dog, Motorbike, People(Person)}. From the VOC2007\_trainval and VOC2012\_trainval, we filtrated the images which include any of the above ten classes of objects to build VOC\_norm\_trainval. VOC\_norm\_test is selected from VOC2007\_test in the same way. The total number of images in VOC\_norm\_trainval, VOC2007\_test and ExDark\_test are 12334, 3760 and 2563, respectively. 

We synthesize the low-light VOC\_dark dataset based on VOC\_norm through the transformation $f(x) = x^\gamma$,
where the value of $\gamma$ is randomly sampled from a uniform distribution within the range $[1.5, 5]$, and $x$ denotes the input pixel intensity. 

\begin{table}[t]
\centering
\scalebox{0.85}{
\begin{tabular}{l|l|l|ccc}

\hline
\multicolumn{2}{c|}{Method}&Train data & V\_n\_ts&V\_D\_t&E\_t\\
\cline{1-6}
  \multirow{2}{*}{Baseline} & YOLOv3 \uppercase\expandafter{\romannumeral1}&VOC\_norm&69.13 & 45.92 &36.42\\ 
  &YOLOv3 \uppercase\expandafter{\romannumeral2}&Hybrid data&65.33 & 52.28 &37.03\\
  \cline{1-3}
  Enhance& ZeroDCE&VOC\_norm&/&33.57 & 34.41 \\ \cline{1-3}
  DA& DAYOLO&Hybrid data&41.68 & 21.53 &18.15\\ \cline{1-3}
  Mul-task&DSNet&Hybrid data&64.08 & 43.75 &36.97\\ \cline{1-3}
  Ours&IA-YOLO&Hybrid data&\textbf{70.02} & \textbf{59.40} &\textbf{40.37}\\ 
  \hline
\end{tabular}
}
\caption{Performance comparison on low-light images. The right three columns list the mAP on the three test datasets, including V\_n\_ts (VOC\_norm\_test), V\_D\_t (VOC\_Dark\_test) and E\_t (Exdark\_test).}
\label{table4}
\end{table}

\subsubsection{Experimental Results} We compare our presented IA-YOLO method with baseline YOLOv3, $Enhance+Detect$, DAYOLO, and DSNet on the three testing datasets. For $Enhanc+Detect$, we employ the recent image enhancement method Zero-DCE~\cite{Zero-DCE} to preprocess the low-light images and use YOLOv3 trained on VOC\_norm for detection. The remaining experimental settings are kept the same as those on foggy images. Table \ref{table4} shows the mAP results. It can be seen that our method yields the best results. IA-YOLO improves the baseline YOLO I by 0.89, 13.48 and 3.95 percent on VOC\_norm\_test, VOC\_Dark\_test and ExDark\_test, respectively, and it improves the YOLO II baseline by 4.69, 7.12 and 3.34 percent on those test sets. This demonstrates that our proposed IA-YOLO approach is also effective in low-light conditions.

Fig.~\ref{fig4} shows the qualitative comparisons between IA-YOLO and the baseline YOLOv3 \uppercase\expandafter{\romannumeral2}. It can be observed that our proposed DIP module is able to adaptively increase the contrast of input image and reveal the image details, which are essential to object detection.



\subsection{Ablation Study}
To examine the effectiveness of each module in our proposed framework, we conduct ablation experiments on different settings, including hybrid data training scheme, DIP and image adaption. We also evaluate the selection of the proposed differentiable filters on three testing datasets.
\begin{figure}[t]
\centering
\includegraphics[width=0.48\textwidth]{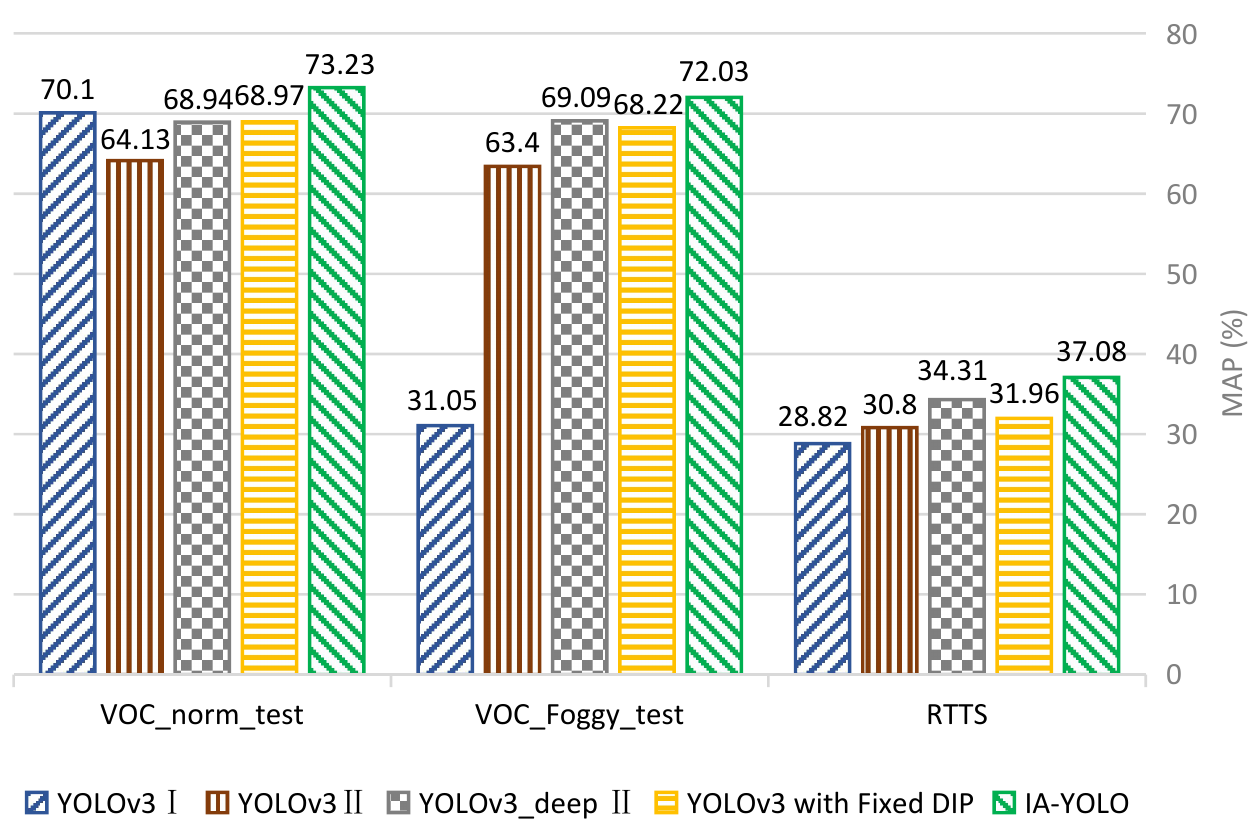} 
\caption{Performance comparison of different settings in foggy conditions.}
\label{fig5}
\end{figure}

The results of the conducted experiments are depicted in Fig.~\ref{fig5}. Except for YOLO I trained with VOC\_norm, the rest of experiments employ the same hybrid data training and the experimental settings. It can be seen that hybrid data training, DIP filter preprocessing, and image adaptive methods all can improve the detection performance on both VOC\_Foggy\_test and RTTS compared with YOLO I. IA-YOLO achieves the best results by using all the three modules. YOLOv3 with fixed DIP means that the filter's hyperparameters are a given set of fixed values, all of which are within a reasonable range.  YOLOv3\_deep \uppercase\expandafter{\romannumeral2} is a deeper version of YOLO \uppercase\expandafter{\romannumeral2} by adding eight convolutional layers with learning parameters over 411K. As shown in Fig.~\ref{fig5}, our proposed IA-YOLO approach performs better than YOLOv3\_deep \uppercase\expandafter{\romannumeral2} with only 165K additional parameters in CNN-PP. It is worthy mentioning that 
only the adaptive learning module improves the performance on VOC\_norm\_test compared against YOLO I in the normal weather conditions, while both YOLOv3 II and YOLOv3 with fixed DIP obtain inferior results. This demonstrates that IA-YOLO can adaptively process both the normal images and foggy ones, which is beneficial for the down-streamed detection task. 

\begin{figure}[t]
\centering
\includegraphics[width=0.48\textwidth]{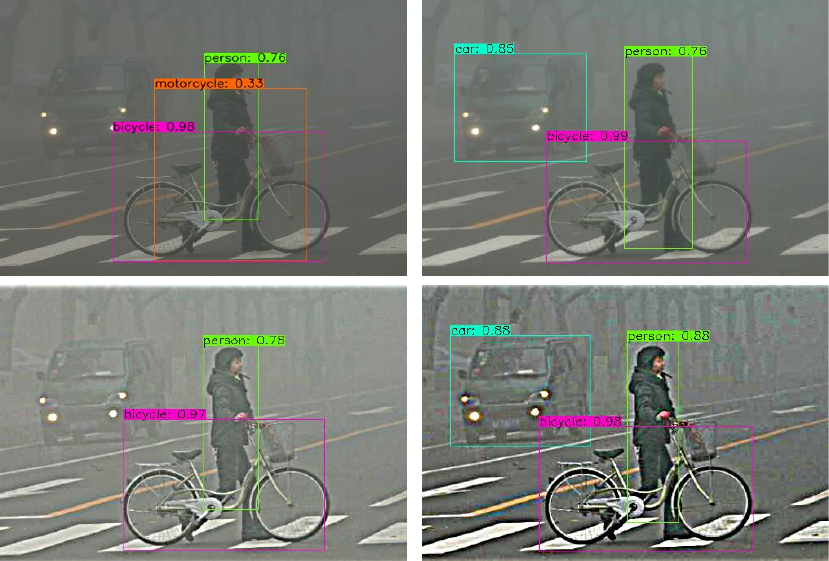} 
\caption{Detection results on RTTS. From left to right, top to bottom: YOLO \uppercase\expandafter{\romannumeral2}, Model B, Model C and Model D.}
\label{fig6}
\end{figure}

As shown in Table \ref{table5}, we conduct quantitative mAP evaluation on filter selection using the three testing datasets. By combining the three sets of filters, \emph{Model D} obtains the best result, demonstrating the effectiveness of these filters. Fig.~\ref{fig6} shows the visual comparisons of several models in Table~\ref{table5}. Compared with \emph{Model C} which brightens and sharpens the image, and \emph{Model B} which defogs the image, the image processed by \emph{Model D} is not only brighter and sharper but also clearer, which makes the foggy objects much more visible. Moreover, we provide some examples on how CNN-PP predicts the DIP module's parameters in the supplementary material. Please refer to the supplementary file for details.

\begin{table}[t]
\centering
\scalebox{0.82}{
\begin{tabular}{c|ccc|ccc}
    \hline
    \multirow{2}{*}{Model}&\multirow{1}{*}{Defog}&\multirow{1}{*}{Pixel\_wise}&\multirow{1}{*}{Sharpen}&\multirow{2}{*}{V\_n\_ts}& \multirow{2}{*}{V\_F\_t}& \multirow{2}{*}{RTTS}\\
    &\multirow{1}{*}{Filter}
    &\multirow{1}{*}{Filters}
    &\multirow{1}{*}{Filter}\\
  \cline{1-7}
  
    A&\cmark&&\cmark&71.47&70.25&34.88\\
    B&\cmark&\cmark&&71.43&70.14&34.83\\
    C&&\cmark&\cmark&70.51&70.09&34.95\\
    D&\cmark&\cmark&\cmark&73.23&72.03&37.08\\

  \hline
\end{tabular}
}
\caption{Ablation analysis on the filters in DIP module.}
\label{table5}
\end{table}
\subsection{Efficiency Analysis}
In our IA-YOLO framework, we introduce a small CNN-PP learning module with 165K trainable parameters into YOLOv3. IA-YOLO takes 44 ms to detect a $544\times544\times3$ resolution image on a single Tesla V100 GPU. It costs only additional 13 ms over the YOLOv3 baseline, while it is 7 ms and 50 ms faster than GridDehaze-YOLOv3 and MSBDN-YOLOv3, respectively. In summary, IA-YOLO only adds 165K trainable parameters while achieving much better performance on all test datasets with comparable running time.

\section{Conclusion}
We proposed a novel IA-YOLO approach to improve object detection in adverse weather conditions, where each input image was adaptively enhanced to obtain better detection performance. A fully differentiable image processing module was developed to restore the latent content by removing the weather-specific information for YOLO detector, whose hyperparameters are predicted by a small convolutional neural network. Moreover, the whole framework was trained in an end-to-end fashion, where the parameter prediction network was weakly supervised to learn an appropriate DIP module through the detection loss. By taking advantages of hybrid training and the parameter prediction network, our proposed approach was able to adaptively handle normal and adverse weather conditions. The experimental results showed that our method performed much better than previous approaches in both the foggy and low-light scenarios.

\section{Acknowledgments}
This work is supported by National Natural Science Foundation of China under Grants (61831015) and Hong Kong RGC RIF Grant (R5001-18). This work is also supported by the funding of ``Leading Innovation Team of the Zhejiang Province" (Grant NO. 2018R01017).
\bibliography{aaai22}

\begin{thebibliography}{39}
\providecommand{\natexlab}[1]{#1}

\bibitem[{Bochkovskiy, Wang, and Liao(2020)}]{bochkovskiy2020yolov4}
Bochkovskiy, A.; Wang, C.-Y.; and Liao, H.-Y.~M. 2020.
\newblock Yolov4: Optimal speed and accuracy of object detection.
\newblock arXiv:2004.10934.

\bibitem[{Chen et~al.(2018)Chen, Li, Sakaridis, Dai, and
  Van~Gool}]{chen2018domain}
Chen, Y.; Li, W.; Sakaridis, C.; Dai, D.; and Van~Gool, L. 2018.
\newblock Domain adaptive faster r-cnn for object detection in the wild.
\newblock In \emph{Proceedings of IEEE/CVF Conference Computer Vision Pattern
  Recognition (CVPR)}, 3339--3348.

\bibitem[{Deng et~al.(2009)Deng, Dong, Socher, Li, Li, and
  Fei-Fei}]{deng2009imagenet}
Deng, J.; Dong, W.; Socher, R.; Li, L.-J.; Li, K.; and Fei-Fei, L. 2009.
\newblock Imagenet: A large-scale hierarchical image database.
\newblock In \emph{Proceedings of IEEE/CVF Conference Computer Vision Pattern
  Recognition (CVPR)}, 248--255. IEEE.

\bibitem[{Everingham et~al.(2010)Everingham, Van~Gool, Williams, Winn, and
  Zisserman}]{everingham2010pascal}
Everingham, M.; Van~Gool, L.; Williams, C.~K.; Winn, J.; and Zisserman, A.
  2010.
\newblock The pascal visual object classes (voc) challenge.
\newblock \emph{International Journal of Computer Vision}, 88(2): 303--338.

\bibitem[{Everingham et~al.(2012)Everingham, Van~Gool, Williams, Winn, and
  Zisserman}]{pascal-voc-2012}
Everingham, M.; Van~Gool, L.; Williams, C. K.~I.; Winn, J.; and Zisserman, A.
  2012.
\newblock The {PASCAL} {V}isual {O}bject {C}lasses {C}hallenge 2012 {(VOC2012)}
  {R}esults.
\newblock
  http://www.pascal-network.org/challenges/VOC/voc2012/workshop/index.html.

\bibitem[{Girshick(2015)}]{girshick2015fast}
Girshick, R. 2015.
\newblock Fast r-cnn.
\newblock In \emph{Proceedings of the IEEE International Conference on Computer
  Vision (ICCV)}, 1440--1448.

\bibitem[{Girshick et~al.(2014)Girshick, Donahue, Darrell, and
  Malik}]{girshick2014rich}
Girshick, R.; Donahue, J.; Darrell, T.; and Malik, J. 2014.
\newblock Rich feature hierarchies for accurate object detection and semantic
  segmentation.
\newblock In \emph{Proceedings of IEEE/CVF Conference Computer Vision Pattern
  Recognition (CVPR)}, 580--587.

\bibitem[{Guo et~al.(2020)Guo, Li, Guo, Loy, Hou, Kwong, and Cong}]{Zero-DCE}
Guo, C.~G.; Li, C.; Guo, J.; Loy, C.~C.; Hou, J.; Kwong, S.; and Cong, R. 2020.
\newblock Zero-reference deep curve estimation for low-light image enhancement.
\newblock In \emph{Proceedings of IEEE/CVF Conference Computer Vision Pattern
  Recognition (CVPR)}, 1780--1789.

\bibitem[{Hang et~al.(2020)Hang, Jinshan, Zhe, Xiang, Xinyi, Fei, and
  Ming-Hsuan}]{MSBDN-DFF}
Hang, D.; Jinshan, P.; Zhe, H.; Xiang, L.; Xinyi, Z.; Fei, W.; and Ming-Hsuan,
  Y. 2020.
\newblock Multi-Scale Boosted Dehazing Network with Dense Feature Fusion.
\newblock In \emph{Proceedings of IEEE/CVF Conference Computer Vision Pattern
  Recognition (CVPR)}.

\bibitem[{He, Sun, and Tang(2009)}]{he2009single}
He, K.; Sun, J.; and Tang, X. 2009.
\newblock Single image haze removal using dark channel prior.
\newblock In \emph{Proceedings of IEEE/CVF Conference Computer Vision Pattern
  Recognition (CVPR)}.

\bibitem[{He et~al.(2016)He, Zhang, Ren, and Sun}]{he2016deep}
He, K.; Zhang, X.; Ren, S.; and Sun, J. 2016.
\newblock Deep residual learning for image recognition.
\newblock In \emph{Proceedings of IEEE/CVF Conference Computer Vision Pattern
  Recognition (CVPR)}, 770--778.

\bibitem[{Hnewa and Radha(2021)}]{hnewa2021multiscale}
Hnewa, M.; and Radha, H. 2021.
\newblock Multiscale Domain Adaptive YOLO for Cross-Domain Object Detection.
\newblock arXiv:2106.01483.

\bibitem[{Hu et~al.(2018)Hu, He, Xu, Wang, and Lin}]{hu2018exposure}
Hu, Y.; He, H.; Xu, C.; Wang, B.; and Lin, S. 2018.
\newblock Exposure: A White-Box Photo Post-Processing Framework.
\newblock \emph{ACM Transactions on Graphics (TOG)}, 37(2): 26.

\bibitem[{Huang, Le, and Jaw(2020)}]{huang2020dsnet}
Huang, S.-C.; Le, T.-H.; and Jaw, D.-W. 2020.
\newblock DSNet: Joint semantic learning for object detection in inclement
  weather conditions.
\newblock \emph{IEEE Transactions on Pattern Analysis and Machine
  Intelligence}.

\bibitem[{Kingma and Ba(2014)}]{kingma2014adam}
Kingma, D.~P.; and Ba, J. 2014.
\newblock Adam: A method for stochastic optimization.
\newblock arXiv:1412.6980.

\bibitem[{Li et~al.(2017)Li, Peng, Wang, Xu, and Feng}]{li2017aod}
Li, B.; Peng, X.; Wang, Z.; Xu, J.; and Feng, D. 2017.
\newblock Aod-net: All-in-one dehazing network.
\newblock In \emph{Proceedings of the IEEE International Conference on Computer
  Vision (ICCV)}, 4770--4778.

\bibitem[{Li et~al.(2018)Li, Ren, Fu, Tao, Feng, Zeng, and
  Wang}]{li2018benchmarking}
Li, B.; Ren, W.; Fu, D.; Tao, D.; Feng, D.; Zeng, W.; and Wang, Z. 2018.
\newblock Benchmarking single-image dehazing and beyond.
\newblock \emph{IEEE Transactions on Image Processing}, 28(1): 492--505.

\bibitem[{Lin et~al.(2014)Lin, Maire, Belongie, Hays, Perona, Ramanan,
  Doll{\'a}r, and Zitnick}]{lin2014microsoft}
Lin, T.-Y.; Maire, M.; Belongie, S.; Hays, J.; Perona, P.; Ramanan, D.;
  Doll{\'a}r, P.; and Zitnick, C.~L. 2014.
\newblock Microsoft coco: Common objects in context.
\newblock In \emph{European Conference on Computer Vision (ECCV)}, 740--755.
  Springer.

\bibitem[{Liu et~al.(2016)Liu, Anguelov, Erhan, Szegedy, Reed, Fu, and
  Berg}]{liu2016ssd}
Liu, W.; Anguelov, D.; Erhan, D.; Szegedy, C.; Reed, S.; Fu, C.-Y.; and Berg,
  A.~C. 2016.
\newblock Ssd: Single shot multibox detector.
\newblock In \emph{European Conference on Computer Vision}, 21--37. Springer.

\bibitem[{Liu et~al.(2019)Liu, Ma, Shi, and Chen}]{liuICCV2019GridDehazeNet}
Liu, X.; Ma, Y.; Shi, Z.; and Chen, J. 2019.
\newblock GridDehazeNet: Attention-Based Multi-Scale Network for Image
  Dehazing.
\newblock In \emph{Proceedings of the IEEE International Conference on Computer
  Vision (ICCV)}.

\bibitem[{Loh and Chan(2019)}]{loh2019getting}
Loh, Y.~P.; and Chan, C.~S. 2019.
\newblock Getting to know low-light images with the exclusively dark dataset.
\newblock \emph{Computer Vision and Image Understanding}, 178: 30--42.

\bibitem[{McCartney(1976)}]{mccartney1976optics}
McCartney, E.~J. 1976.
\newblock Optics of the atmosphere: scattering by molecules and particles.
\newblock \emph{New York}.

\bibitem[{Mosleh et~al.(2020)Mosleh, Sharma, Onzon, Mannan, Robidoux, and
  Heide}]{mosleh2020hardware}
Mosleh, A.; Sharma, A.; Onzon, E.; Mannan, F.; Robidoux, N.; and Heide, F.
  2020.
\newblock Hardware-in-the-loop end-to-end optimization of camera image
  processing pipelines.
\newblock In \emph{Proceedings of the IEEE/CVF Conference on Computer Vision
  and Pattern Recognition (CVPR)}, 7529--7538.

\bibitem[{Narasimhan and Nayar(2002)}]{narasimhan2002vision}
Narasimhan, S.~G.; and Nayar, S.~K. 2002.
\newblock Vision and the atmosphere.
\newblock \emph{International Journal of Computer Vision}, 48(3): 233--254.

\bibitem[{Polesel, Ramponi, and Mathews(2000)}]{polesel2000image}
Polesel, A.; Ramponi, G.; and Mathews, V.~J. 2000.
\newblock Image enhancement via adaptive unsharp masking.
\newblock \emph{IEEE Transactions on Image Processing}, 9(3): 505--510.

\bibitem[{Qin et~al.(2020)Qin, Wang, Bai, Xie, and Jia}]{qin2020ffa}
Qin, X.; Wang, Z.; Bai, Y.; Xie, X.; and Jia, H. 2020.
\newblock FFA-Net: Feature fusion attention network for single image dehazing.
\newblock In \emph{Proceedings of the AAAI Conference on Artificial
  Intelligence}, volume~34, 11908--11915.

\bibitem[{Redmon et~al.(2016)Redmon, Divvala, Girshick, and
  Farhadi}]{redmon2016you}
Redmon, J.; Divvala, S.; Girshick, R.; and Farhadi, A. 2016.
\newblock You only look once: Unified, real-time object detection.
\newblock In \emph{Proceedings of IEEE/CVF Conference Computer Vision Pattern
  Recognition (CVPR)}, 779--788.

\bibitem[{Redmon and Farhadi(2017)}]{redmon2017yolo9000}
Redmon, J.; and Farhadi, A. 2017.
\newblock YOLO9000: better, faster, stronger.
\newblock In \emph{Proceedings of IEEE/CVF Conference Computer Vision Pattern
  Recognition (CVPR)}, 7263--7271.

\bibitem[{Redmon and Farhadi(2018)}]{redmon2018yolov3}
Redmon, J.; and Farhadi, A. 2018.
\newblock Yolov3: An incremental improvement.
\newblock arXiv:1804.02767.

\bibitem[{Ren et~al.(2015)Ren, He, Girshick, and Sun}]{ren2015faster}
Ren, S.; He, K.; Girshick, R.; and Sun, J. 2015.
\newblock Faster r-cnn: Towards real-time object detection with region proposal
  networks.
\newblock \emph{Advances in Neural Information Processing Systems}, 28: 91--99.

\bibitem[{Sindagi et~al.(2020)Sindagi, Oza, Yasarla, and
  Patel}]{sindagi2020prior}
Sindagi, V.~A.; Oza, P.; Yasarla, R.; and Patel, V.~M. 2020.
\newblock Prior-based domain adaptive object detection for hazy and rainy
  conditions.
\newblock In \emph{European Conference on Computer Vision (ECCV)}, 763--780.
  Springer.

\bibitem[{Wang et~al.(2019)Wang, Guo, Chen, Li, and Xu}]{wang2019pso}
Wang, G.; Guo, J.; Chen, Y.; Li, Y.; and Xu, Q. 2019.
\newblock A PSO and BFO-based learning strategy applied to faster R-CNN for
  object detection in autonomous driving.
\newblock \emph{IEEE Access}, 7: 18840--18859.

\bibitem[{Wang et~al.(2021)Wang, Chen, Yuan, and Guan}]{wang2021adaptive}
Wang, W.; Chen, Z.; Yuan, X.; and Guan, F. 2021.
\newblock An adaptive weak light image enhancement method.
\newblock In \emph{Twelfth International Conference on Signal Processing
  Systems}, volume 11719, 1171902. International Society for Optics and
  Photonics.

\bibitem[{You et~al.(2015)You, Tan, Kawakami, Mukaigawa, and
  Ikeuchi}]{you2015adherent}
You, S.; Tan, R.~T.; Kawakami, R.; Mukaigawa, Y.; and Ikeuchi, K. 2015.
\newblock Adherent raindrop modeling, detectionand removal in video.
\newblock \emph{IEEE Transactions on Pattern Analysis and Machine
  Intelligence}, 38(9): 1721--1733.

\bibitem[{Yu et~al.(2018)Yu, Liu, Zhang, Qu, Zhao, and
  Zhang}]{yu2018deepexposure}
Yu, R.; Liu, W.; Zhang, Y.; Qu, Z.; Zhao, D.; and Zhang, B. 2018.
\newblock Deepexposure: Learning to expose photos with asynchronously
  reinforced adversarial learning.
\newblock In \emph{Proceedings of the 32nd International Conference on Neural
  Information Processing Systems (NeurIPS)}, 2153--2163.

\bibitem[{Yu and Bajaj(2004)}]{yu2004fast}
Yu, Z.; and Bajaj, C. 2004.
\newblock A fast and adaptive method for image contrast enhancement.
\newblock In \emph{2004 International Conference on Image Processing, 2004.
  ICIP'04.}, volume~2, 1001--1004. IEEE.

\bibitem[{Zeng et~al.(2020)Zeng, Cai, Li, Cao, and Zhang}]{zeng2020learning}
Zeng, H.; Cai, J.; Li, L.; Cao, Z.; and Zhang, L. 2020.
\newblock Learning image-adaptive 3D lookup tables for high performance photo
  enhancement in real-time.
\newblock \emph{IEEE Transactions on Pattern Analysis and Machine
  Intelligence}.

\bibitem[{Zhang et~al.(2021)Zhang, Tuo, Hu, and Jing}]{zhang2021domain}
Zhang, S.; Tuo, H.; Hu, J.; and Jing, Z. 2021.
\newblock Domain Adaptive YOLO for One-Stage Cross-Domain Detection.
\newblock arXiv:2106.13939.

\bibitem[{Zhao et~al.(2019)Zhao, Zheng, Xu, and Wu}]{zhao2019object}
Zhao, Z.-Q.; Zheng, P.; Xu, S.-t.; and Wu, X. 2019.
\newblock Object detection with deep learning: A review.
\newblock \emph{IEEE Transactions on Neural Networks and Learning Systems},
  30(11): 3212--3232.

\end{thebibliography}

\clearpage
\section{Supplementary Material}
\subsection{Defog Filter Design}

Motivated by the conventional dark channel prior method~\cite{he2009single}, we design a defog filter with a learnable parameter. In the atmospheric
scattering model~\citep{mccartney1976optics,narasimhan2002vision}, the formation of a hazy image can be formulated as follows:
\begin{equation}
I(x) = J(x)t(x) + A(1-t(x))\label{12}
\end{equation} where $I(x)$ is the foggy image, and $J(x)$ represents the scene radiance (clear image). $A$ is the global atmospheric light, and $t(x)$ is the medium transmission map.

In order to recover the clear image $J(x)$, the key is to obtain the global atmospheric light $A$ and the medium transmission map $t(x)$. To this end, we first compute the dark channel map and pick the top 1000 brightest pixels. Then, $A$ is estimated by averaging these 1000 pixels in the haze image $I(x)$. 
From Eq. (\ref{12}), we can derive that
\begin{equation}
\frac{I^C (x)}{A^C}= t(x)\frac{J^C (x)}{A^C} + (1-t(x))\label{13}
\end{equation}
where $C$ donates the RGB color channel. By taking two min operations, one on the channels and one on a local patch, in the above equation, we can obtain:
\begin{equation}
\min \limits_{C}( \min \limits_{y \in \Omega(x)} \frac{I^C (y)}{A^C} )= t(x)\min \limits_{C}( \min \limits_{y \in \Omega(x)} \frac{J^C (y)}{A^C} ) + (1-t(x))\label{14}
\end{equation}

Based on the dark channel prior, we can get that
\begin{equation}
J ^ {dark} \ (x) = \min \limits_{C}( \min \limits_{y \in \Omega(x)} J^C (y) ) =0 \label{15}
\end{equation}
Since $A^C$ is always positive, Eq. (\ref{15}) can be written as:
\begin{equation}
\min \limits_{C} (\min \limits_{y \in \Omega(x)} \frac{J^C (y)}{A^C}) = 0 \label{16}
\end{equation}
By substituting Eq. (\ref{16}) into Eq. (\ref{13}), we can obtain:
\begin{equation}
t(x) = 1-  \min \limits_{C} (\min \limits_{y \in \Omega(x)} \frac{I^C (y)}{A^C}) \label{17}
\end{equation}

We further introduce a parameter $\omega$ to control the degree of defogging. There is:
\begin{equation}
t(x, \omega) = 1-  \omega \min \limits_{C} (\min \limits_{y \in \Omega(x)} \frac{I^C (y)}{A^C}) \label{18}
\end{equation}
Since the above operation is differentiable, we can optimize $\omega$ through back propagation to make \emph{defog filter} more conducive to foggy image detection.

\subsection{Experiments}
\subsubsection{Experiments on Foggy Images}

We compare our method with the baseline YOLOv3~\cite{redmon2018yolov3}, $Defog+Detect$~\citep{MSBDN-DFF, liuICCV2019GridDehazeNet}, domain adaptation~\cite{hnewa2021multiscale}, and multi-task learning~\cite{huang2020dsnet}. For domain adaptation approach, we employ the one-stage multi-scale domain-adaptive detector DAYOLO~\cite{hnewa2021multiscale} with multiple domain adaptation paths and the corresponding domain classifiers at different scales of YOLOv3. We set the loss weight $\lambda = 0.1$ for training, and each batch has 2 images, one from the source domain and the other from the target domain. Other hyperparameters are set the same as in the original paper.

\begin{figure*}
\centering
\includegraphics[width=1\textwidth]{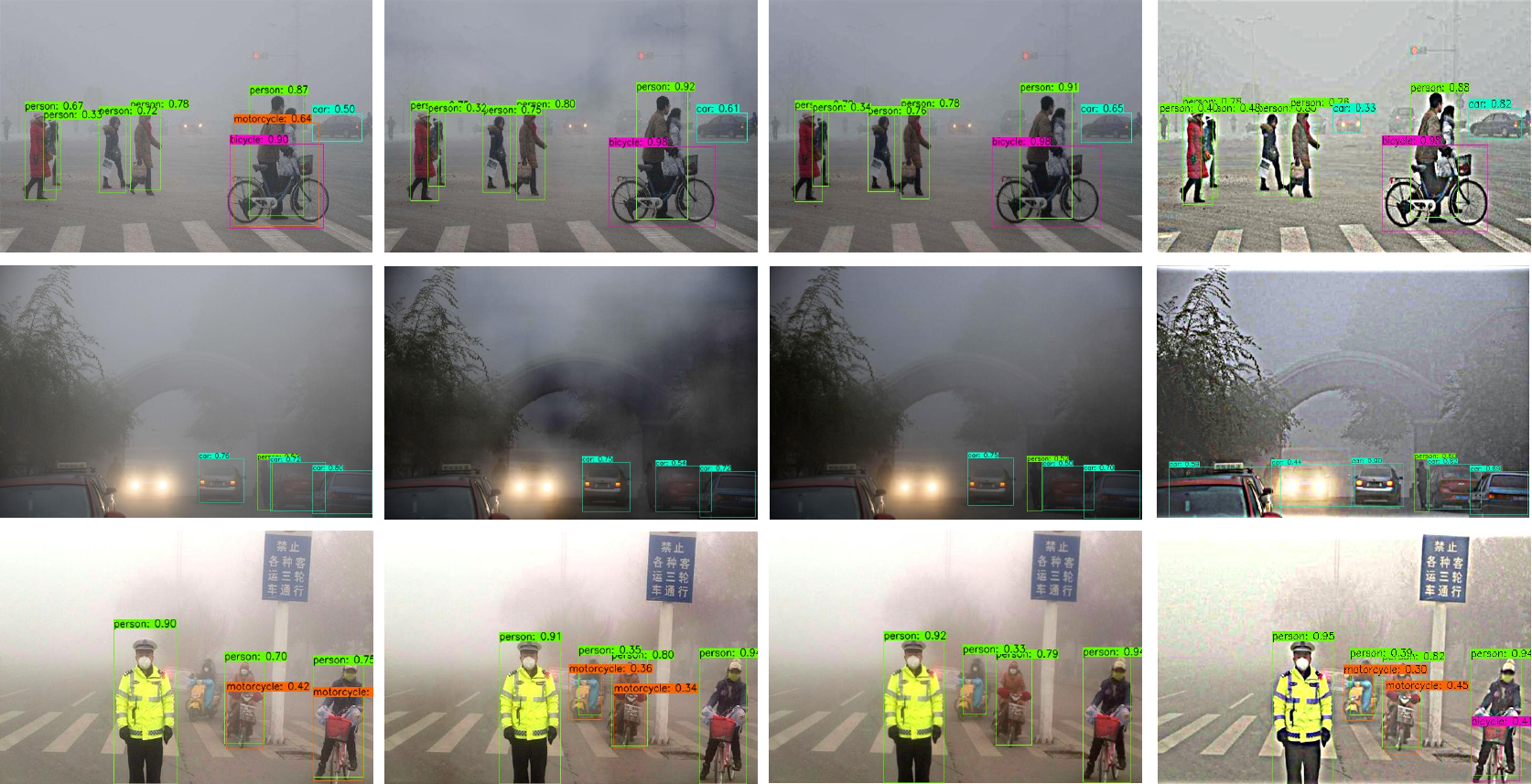} 
\caption{Detection results by different methods on real-world RTTS foggy images. From left to right: YOLOv3 \uppercase\expandafter{\romannumeral2}, GirdDehaze + YOLOv3 \uppercase\expandafter{\romannumeral1}, MSBDN + YOLOv3 \uppercase\expandafter{\romannumeral1} and our IA-YOLO.  The proposed method learns to reduce the haze and enhance the image contrast, which leads to better detection performance with fewer missed and wrong detections.}
\label{fig7}
\end{figure*}
 
\begin{table*}[t]
\centering
\scalebox{1.0}{
\begin{tabular}{lcccccccccccc}
    \toprule
    Dataset & person & bicycle & car & bus & motorbike& boat & bottle & cat & chair & dog& Total \\
    \midrule
    Voc\_norm\_trainval &  13256 & 1064 & 3267 & 822 & 1052 & 1140 & 1764 & 1593 & 3152 & 2025 & 29135 \\
    Voc\_norm\_test  &  4528 & 337 & 1201 & 213 & 325& 263 & 469 & 358 & 756 & 489 &8939 \\
    ExDark\_test & 2235 &  418 & 919 & 164 & 242 &515 &  433 & 425 & 609 & 490& 6450 \\
    \bottomrule
\end{tabular}
}
\caption{Statistics of the used datasets.}
\label{table6}
\end{table*}
 
Fig.~\ref{fig7} shows several visual examples of our IA-YOLO method, the baseline YOLOv3 \uppercase\expandafter{\romannumeral2} and the $Defog+Detect$ methods. Both GridDehaze~\cite{liuICCV2019GridDehazeNet} and MSBDN~\cite{MSBDN-DFF} can reduce the haze effect, which is generally beneficial to detection. Our IA-YOLO method not only reduces the haze, but also enhances the local image gradients, which lead to better detection performance.

\begin{figure*}
\centering
\includegraphics[width=0.95\textwidth]{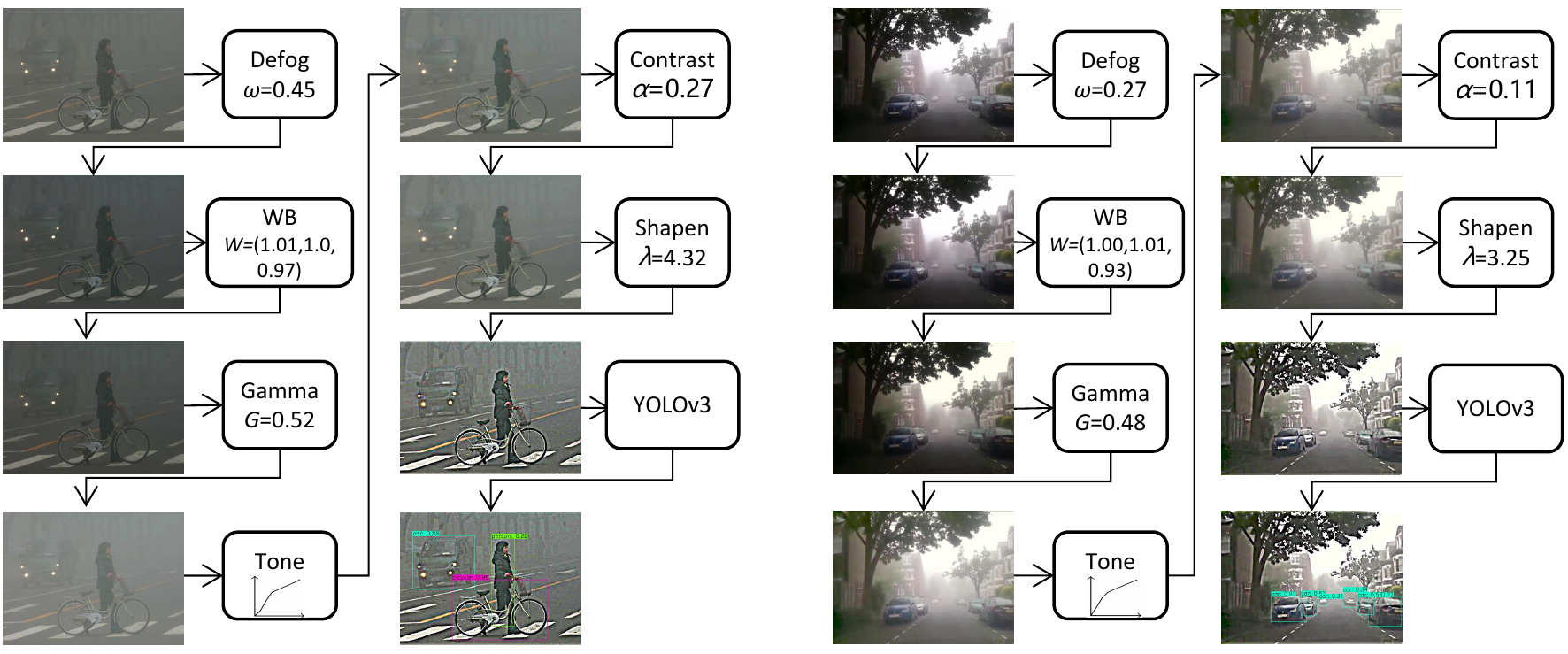} 
\caption{Examples of learned DIP module and their filtering outputs. The image-adaptive processing module can output the corresponding filter parameters according to the brightness, color, tone and weather-information of each input image, so as to get better detection performance.}
\label{fig8}
\end{figure*}

Fig.~\ref{fig8} shows two examples on how the CNN-PP module predicts DIP's parameters , including detailed parameter values and the images processed by each sub-filter. The CNN-PP is able to learn a set of DIP parameters for each image according to its brightness, color, tone and weather-specific information. After the input image is processed by the learned DIP module, more image details are revealed, which are conducive to the subsequent detection task.

\subsubsection{Experiments on Low-light Images}
The total number of images in VOC\_norm\_trainval, VOC\_norm\_test and ExDark\_test are 12334, 3760 and 2563, respectively. The numbers of instances are listed in Table~\ref{table6}.

\begin{figure*}
\centering
\includegraphics[width=0.77\textwidth]{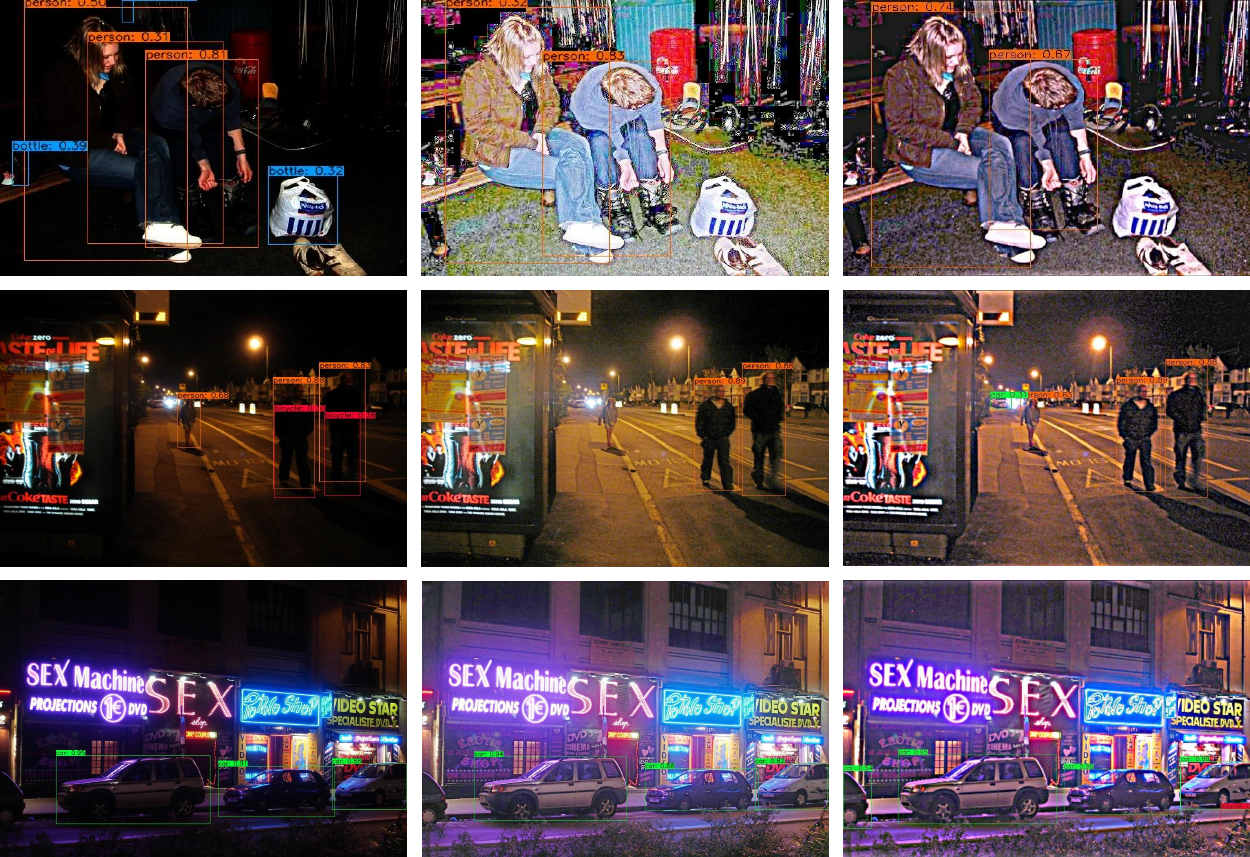} 
\caption{Detection results of different methods on synthetic VOC\_Dark\_test images (top row), real-world ExDark\_test low-light images (bottom two rows). From left to right: YOLOv3 \uppercase\expandafter{\romannumeral2}, ZeroDCE + YOLOv3 \uppercase\expandafter{\romannumeral1} and our IA-YOLO. The proposed method learns to make the image brighter with more details, which results in better detection performance with fewer missed and wrong detections.}
\label{fig9}
\end{figure*}

 We compare our presented method with the baseline YOLOv3, $Enhance+Detect$~\cite{Zero-DCE}, DAYOLO, and DSNet on the three testing datasets. Fig.~\ref{fig9} shows several visual examples of our IA-YOLO method, the baseline YOLOv3 \uppercase\expandafter{\romannumeral2} and the $Enhance+Detect$ methods. It can be observed that both Zero-DCE~\cite{Zero-DCE} and IA-YOLO can brighten the image and reveal the image details. The proposed IA-YOLO can further increases the contrast of the input image, which is essential to object detection.
 
\begin{table}[t]
\centering
\scalebox{1.0}{
\begin{tabular}{ccc}
    \hline
  
    Method&Additional Params&Speed(ms)\\
    \cline{1-3}
    YOLOv3 &/&31\\
    YOLOv3\_deep \uppercase\expandafter{\romannumeral2}&412K&35\\
    ZeroDCE&79K&34\\
    MSBDN&31M&94\\
    GridDehaze&958K&51\\
    IA-YOLO(Ours)&165K&44\\

  \hline
\end{tabular}
}
\caption{Efficiency analysis on the compared methods.}
\label{table7}
\end{table}

\subsection{Efficiency Analysis}
In our proposed IA-YOLO framework, we introduce a learning module of CNN-PP into YOLOv3, which is a small network containing five convolutional layers and two fully connected layers. Table \ref{table7} shows the efficiency analysis of some methods used in our experiments. The methods not listed are validated using the YOLOv3 architecture. The second column lists the number of additional parameters over the YOLOv3 model. The third column lists the running time on a $544\times544\times3$ resolution image with a single Tesla V100 GPU. It can be seen that IA-YOLO only adds 165K trainable parameters over YOLOv3 while achieving the best performance on all testing with comparable running time. Note that IA-YOLO has fewer trainable parameters than YOLO\_deep \uppercase\expandafter{\romannumeral2} but its running time is longer. This is because that the filtering process in the DIP module incurs additional computation.




\end{document}